\documentclass[conference]{IEEEtran}
\IEEEoverridecommandlockouts
\usepackage{cite}
\usepackage{amsmath,amssymb,amsfonts}
\usepackage{algorithmic}
\usepackage{graphicx}
\usepackage{textcomp}
\usepackage{xcolor}
\usepackage{subfigure}
\usepackage{siunitx}
\usepackage{mhchem}
\usepackage{comment}
\usepackage{bm}
\usepackage{adjustbox}
\usepackage{tabularx}
\usepackage{multirow}
\usepackage{dblfloatfix} 
\usepackage[hyphens]{url}

\usepackage{bm}
\def\BibTeX{{\rm B\kern-.05em{\sc i\kern-.025em b}\kern-.08em
    T\kern-.1667em\lower.7ex\hbox{E}\kern-.125emX}}
\begin{document}

\title{Constructing Cloze Questions Generatively

}
\author{\IEEEauthorblockN{1\textsuperscript{st}Yicheng Sun}
\IEEEauthorblockA{\textit{Miner School of Computer and Information Sciences} \\
\textit{University of Massachusetts Lowell}\\
Lowell, USA \\
yicheng\_sun@student.uml.edu}
\and
\IEEEauthorblockN{2\textsuperscript{nd} Jie Wang}
\IEEEauthorblockA{\textit{Miner School of Computer and Information Sciences} \\
\textit{University of Massachusetts Lowell}\\
Lowell, USA \\
jie\_wang@uml.edu}

}

\maketitle

\begin{abstract}
We present a generative method called CQG for constructing cloze questions from a given article using neural networks and WordNet, with an emphasis on generating multigram distractors. Built on sense disambiguation, text-to-text transformation,
WordNet's synset taxonomies and lexical labels, 
CQG selects an answer key for a given sentence,
segments it into a sequence of instances, generates instance-level distractor candidates (IDCs) using a transformer and
sibling synsets. 
It then removes inappropriate IDCs, ranks the remaining IDCs based on contextual embedding similarities, as well as synset and lexical relatedness, forms distractor candidates by combinatorially replacing instances with the
corresponding top-ranked IDCs, and checks if they are legitimate phrases. Finally, it selects top-ranked distractor candidates  based on contextual semantic similarities to the answer key.
Experiments show that this method significantly outperforms SOTA results. Human judges also confirm the high qualities of the generated distractors.
\end{abstract}


\section{Introduction}
Cloze questions are a common mean for assessing reading comprehension. A standard cloze question consists of a stem -- a sentence with one or more blanks -- and four answer
choices for each blank, with one choice being
the correct answer (aka. the answer key) and the rest being the distractors.
Appropriate distractors must be reliable and plausible. 
A distractor is reliable if filling in the blank with it 
yields a contextually fit and grammatically correct yet logically wrong sentence. 
A distractor is plausible if it is not manifestly incorrect, namely, 
if it presents sufficient confusion as to which is the correct answer.

Mining WordNet \cite{miller1998wordnet} 
is a common approach to finding distractors 
that share the same hypernym of the answer key. WordNet offers exquisite taxonomies of English nouns, verbs, adjectives, and adverbs, as well as noun and verb phrases.
This approach, however, could easily lead to inappropriate distractors.
For example, suppose that the word \textsl{dog} is the answer key selected from a sentence about domestic animals. In WordNet, the word \textsl{dog} in the sense of animal has \textsl{canine} as a hypernym, with hyponyms \textsl{wolf}, \textsl{fox}, \textsl{wild dog}, and others. Selecting any of these hyponyms as distractors is inappropriate for the context.

We devise a generative method called CQG (Cloze Question Generator) to generate cloze questions
using text-to-text transformer neural nets and WordNet. On a given article,
CQG carries out the following tasks:
select declarative sentences as stems according to their relative importance,
select answer keys, segment an answer key into instances, determine each instance's lexical label with respect to the underlying synset after sense disambiguation, 
generate instance-level distractor candidates (IDCs) using a transformer and the
hypernym-hyponym hierarchical structure, filter inappropriate IDCs using a number of syntactic and semantic indicators,
rank the remaining IDCs according to similarities of contextual embeddings and synsets with the same
lexical label, and combinatorially substitute top-ranked IDCs for the corresponding
instance in the answer key to form new phrases.
We filter these phrases to produce a pool of distractor candidates that conform to human writings. Finally, we select from the pool a fixed number of distractors that are contextually and semantically closest to the answer key.

Experiments show that CQG significantly outperforms SOTA results \cite{ren2021knowledge} on unigram answer keys. 
For multigram answer keys and distractors, we note that the standard token-frequency-based measures are ill-suited to measuring multigram distractors, and there are no existing results to compare with.
To overcome this obstacle,
we construct a dataset with multigram answer keys in the same way
as the dataset with unigram answer keys is constructed, and show that
CQG generates distractors with over 81\% contextual semantic similarity with the ground truth distractors.
Human judges also confirm the high qualities of both unigram and multigram distractors generated by CQG. 

The major contribution of this paper is the construction of the Cloze Question Generator (CQG), built on sentence ranking, text-to-text transformers, sense disambiguation, and WordNet's vocabulary synsets and lexical labels. CQG enhances contextual understanding, generates distractors of high quality, manages effectively and multigram answer keys, and produces more appropriate cloze questions.

The rest of the paper is organized as follows:  We summarize in Section \ref{sec:related-work} related works.
We describe 
in Section \ref{sec:answer-keys} the architecture of CQG, the selections of stems and answer keys, and the segmentation of multigram answer keys into a sequence of instances.
We present in Section \ref{sec:candidates} how to generate instance-level distractors, and
describe in Section \ref{sec:distractors}
how to generate distractors.
We then present in Section \ref{sec:evaluations} 
datasets, experiments, and evaluation results. Section \ref{sec:conclusions} concludes the paper with final remarks.

\section{Related Works}\label{sec:related-work}
Research on finding distractors is along the following line: Find candidates based on answer keys and the underlying sentences, and select distractors from candidates according to certain metrics. 
Finding reasonable distractor candidates is challenging.  
A common approach extracts words and phrases that can serve as answer keys from (possibly domain-specific) vocabularies, thesauri, and
taxonomies \cite{sumita2005measuring, mitkov2009semantic, smith2010gap, jiang2017distractor, ren2021knowledge},  
and for each answer key ranks the rest of the words and phrases
according to certain criteria.
These are extractive methods that search for distractor candidates from a text corpus or a knowledge base.

WordNet is a large and growing knowledge base of Eng\-lish nouns, verbs, adjectives, and adverbs, as well as
phrases, where words and phrases are grouped
into sets of cognitive synonyms called synsets, with each synset expressing a distinct concept.
Synsets are indexed and interconnected. 
A synset's parent node is called a hypernym synset (where there is no confusion, the word synset
may be omitted) and its siblings are hyponyms. 
The relations of hypernyms and hyponyms of synsets 
are particularly useful for finding distractors. 
Probase \cite{wu2012probase} that supports semantic search is another knowledge base useful for finding distractors for a given answer key.

Intuitively, 
an entry item in WordNet could be used as an answer key as well as a distractor for a different answer key,
 and entry items in a synset with the same hypernym of the answer key's synset are distractor candidates. 
This straightforward application of hypernym-hyponym structure may lead to 
distractors of wrong senses, and the following remedies have been attempted:
(1) Use topic distributions such as those generated by the LDA topic model \cite{blei2003latent} to determine the sense of the answer key \cite{ren2021knowledge} and then
use context search to select candidates with the corresponding sense \cite{susanti2018automatic}. 
WordNet CSG+DS and Probase CSG+DS \cite{ren2021knowledge}, for example, are such attempts, 
where the former uses hypernyms and the latter uses is\_a\_relation to find candidates.
(2) Identify distractors directly from the underlying text through visual and informal examination \cite{knoop2013wordgap}.

Distractors are selected from candidates  
according to certain metrics of similarities to the answer key, including embedding-vector similarities \cite{guo2016questimator} 
difficulty levels \cite{brown2005automatic}, 
WordNet-based metrics \cite{mitkov2003computer}, 
 and syntactic features \cite{agarwal2011automatic}. 
Researchers also considered semantic relatedness of distractors with the whole stem and 
domain restrictions \cite{pino2008selection,mostow2012generating}, and 
 explored machine-learning ranking models to select distractors  and quantitatively evaluate the top generated distractors \cite{liang2017distractor,liang2018distractor,ren2021knowledge}. 

Despite extensive efforts, the qualities of the
distractors produced by existing methods are still
below satisfactory. We present a generative method to improve satisfactions.

\section{Selections of Stems and Answer Keys} 
\label{sec:answer-keys}

\begin{figure}[htb] 
	\centering 
	\includegraphics[width=\columnwidth]{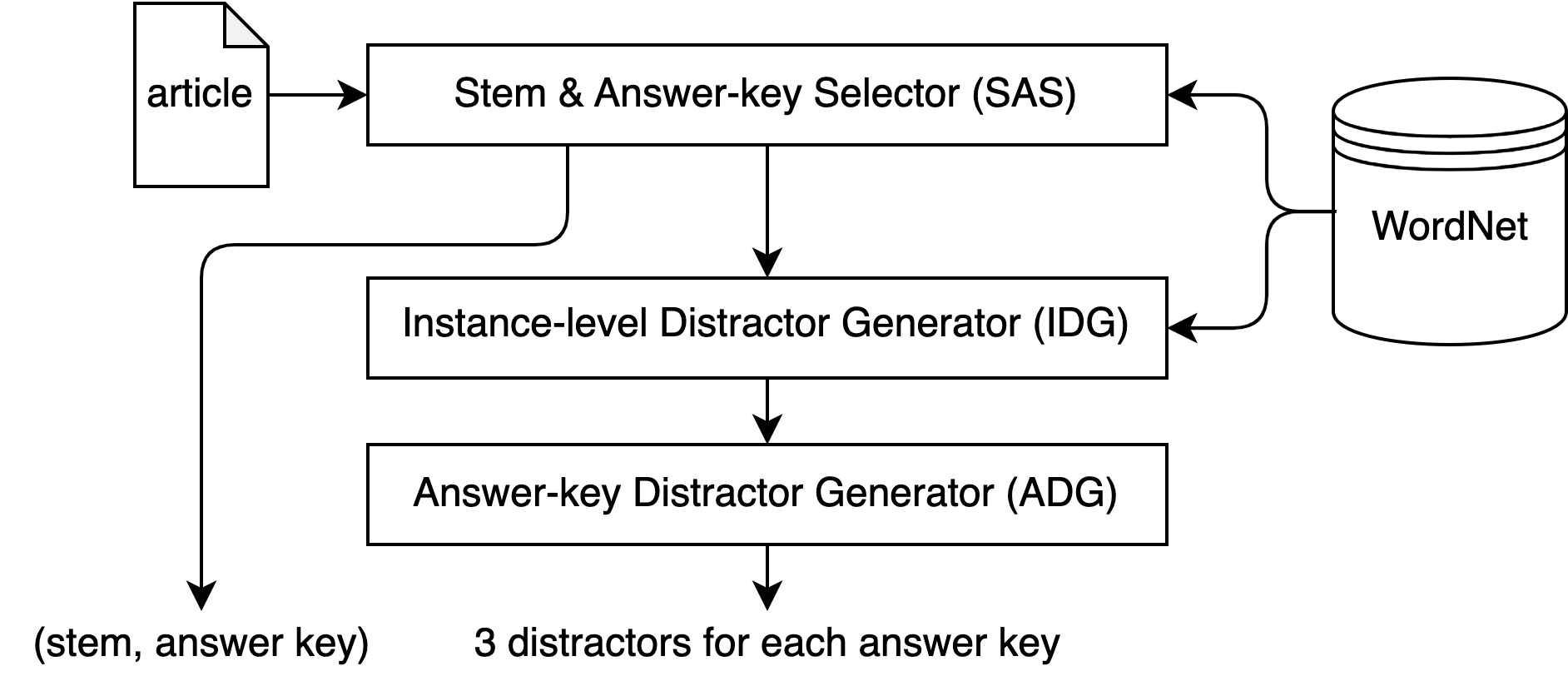} 
	\caption{CQG architecture and data flow} 
	\label{fig:architecture}
\end{figure}

CQG consists of three components as shown in Fig. \ref{fig:architecture}: (1) Stem and Answer-key Selector (SAS). (2) Instance-level Distractor Generator (IDG). (3) Answer-key Distractor Generator (ADG). 

\begin{figure}[bbp]
\centering
\includegraphics[width=\columnwidth]{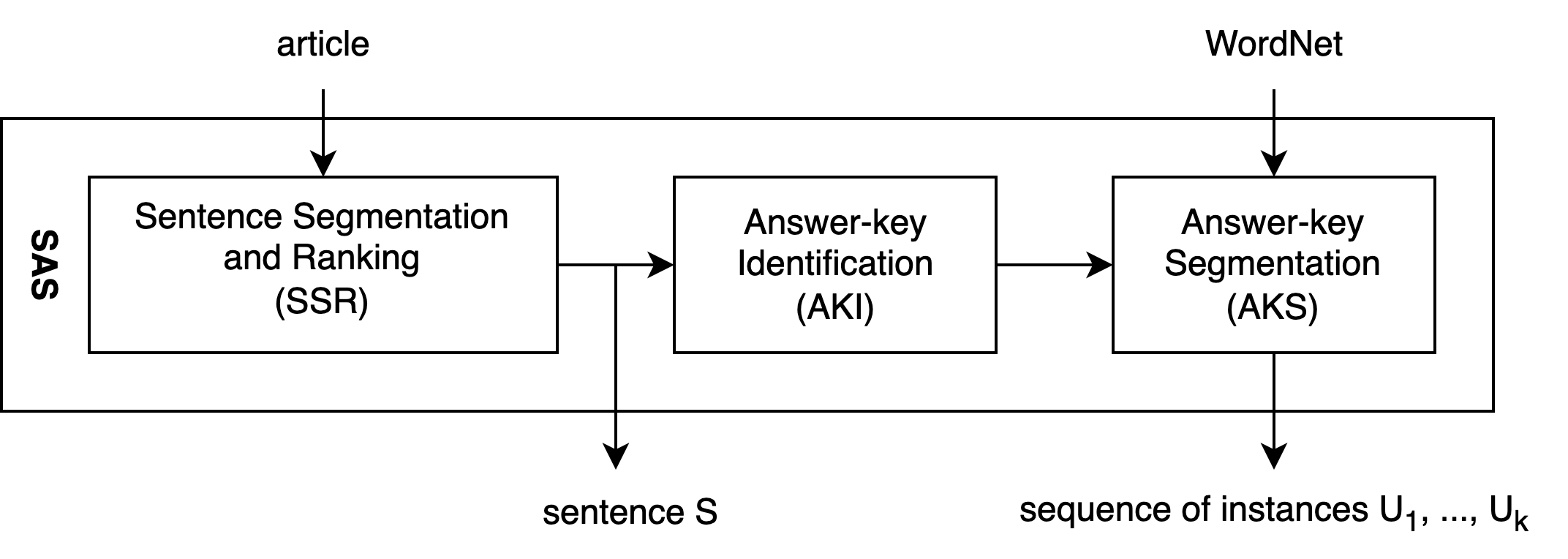}
\caption{SAS sub-components and data flow}
\label{fig:SAKS}
\end{figure}

SAS consists of three sub-components (see Fig. \ref{fig:SAKS}): 
(1) Sentence Segmentation and Ranking (SSR). (2) Answer-key
Identification
(AKI). (3) Answer-key
Segmentation
(AKS).

SSR 
segments a given article into sentences using an NLP tool such as spaCy \cite{spacy},
ranks sentences according to their importance with respect to the article

using an algorithm named Contextual Network and Topic Analysis Rank \cite{Zhang-Zhou-Wang2021},
and selects declarative sentences of reasonable lengths as stems according to their ranks depending on how many cloze questions are to be generated.

AKI selects an answer key from a stem.  A noun or a verb (unigram), and a noun chunk or a verb chunk (multigram) that do not include pronouns in a declarative sentence may be selected as an answer key. Finally, if the selected answer key is a multigram, then SAS segments it into
a sequence of instances, where an instance in a multigram answer key is
the largest legitimate entry in WordNet. 
For example, in the answer key \textsl{abnormal white blood cells}, the word \textsl{cells} or the phrase
\textsl{blood cells} are not instances even though they are both legitimate entries in WordNet,
because \textsl{white blood cells} is a legitimate entry in WordNet. On the other hand,
\textsl{white blood cells} is an instance for 
\textsl{abnormal white blood cells} is not a legitimate entry in WordNet. 
%
%
%
Answer keys can be identified using spaCy's syntactic dependency parser or other similar NLP tools. A POS tagger is used to identify whether an answer key is a noun or a verb chunk. A noun chunk 
is a non-recursive structure consisting of a head noun with zero or more premodifying adjectives and nouns. 
For example, the sentence ``\textsl{The big red apple fell on the scared cat"} consists of two noun chunks:
\textsl{the big red apple} and \textsl{the scared cat}. 



AKS divides a multigram answer key into a sequence of instances, where an instance may be a single word or a sequence of words, and an answer key may consist of one or more instances.
%
%
%
%
Let $X$ be a noun chunk identified as an answer key (verb chunks as an answer key are handled similarly). 
Denote $X$'s suffix of length
$\ell$ by
$X_s[\ell]$. 
We determine instances in the order of right-to-left as follows:
Initially set $\ell \leftarrow 0$. If querying $X_s[\ell+1]$ to WordNet returns a result, then set $\ell \leftarrow \ell+ 1$. Repeat the same procedure until WordNet does not return a result on $X_s[\ell+1]$
or $\ell+1 > |X|$ (the length of $X$). Then $X_s[\ell]$
is an instance. Remove $X_s[\ell]$ from $X$. If the new string is non-empty then repeat the same procedure on the new string.

\begin{figure}[tbp]
\centering
\includegraphics[width=\columnwidth]{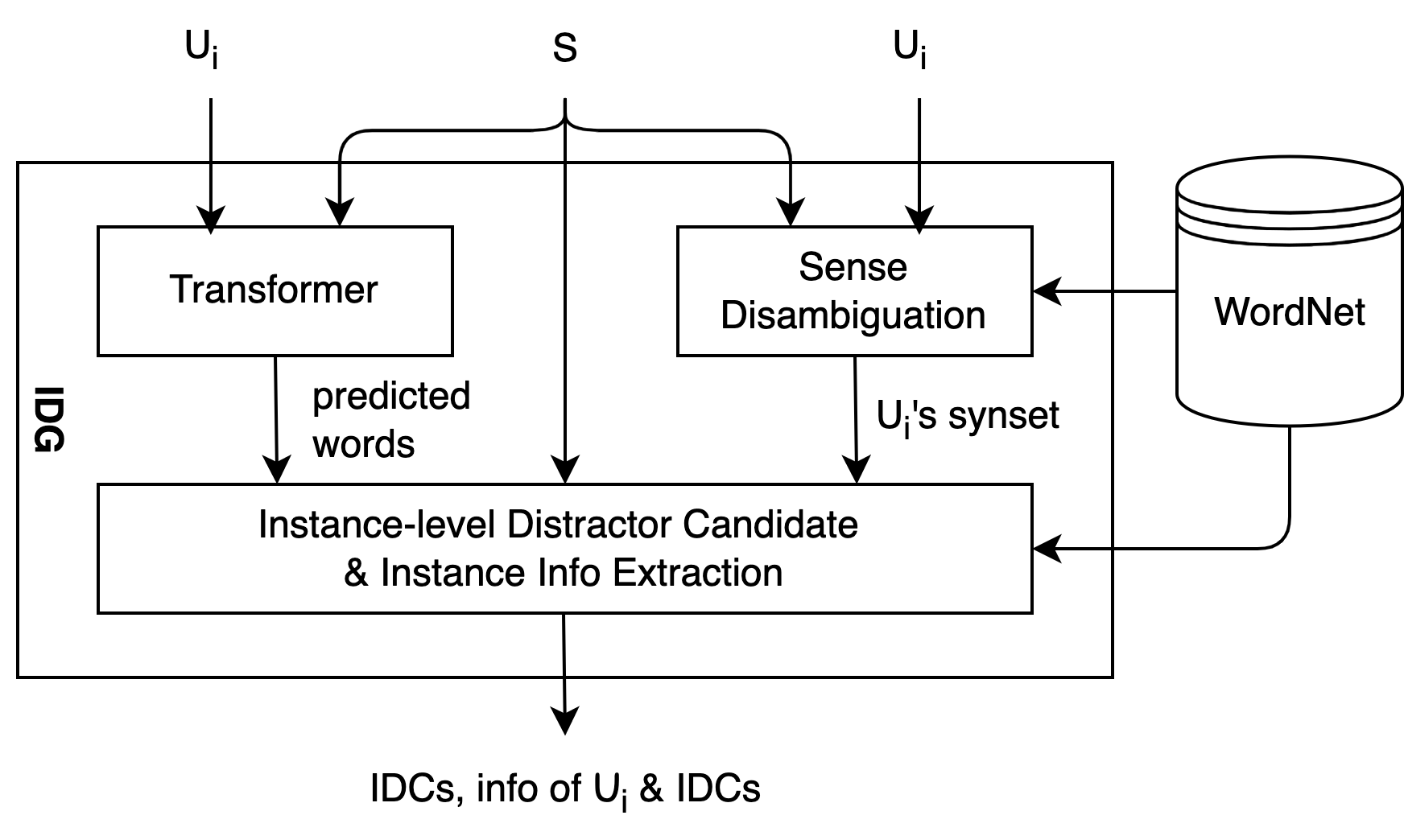}
\caption{IDG sub-components and data flow}
\label{fig:IDG}
\end{figure}

\textbf{Remark}. If we change the search order, then it may lead to instances of different senses.
For example, spaCy identifies \textsl{artificial blood cells} as a noun chunk in a sentence, which is not an entry in WordNet. Using our algorithm we identify \textsl{blood cells} as one instance and \textsl{artificial} the other instance.
If, however, we change the parsing order to left-to-right, we will get two instances \textsl{artificial blood} and \textsl{cells}. To avoid ambiguity, it is necessary
to fix 
a parsing order and for the English language, it makes more sense 
to follow the order of right-to-left.

\section{Instance-level Distractor Generator} 
\label{sec:candidates}

IDG consists of three sub-components (see Fig. \ref{fig:IDG}): (1) Transformer. (2) Sense Disambiguation. (3)
Instance-level Distractor Candidate \& Instance Info Extraction.

\paragraph{Transformer}

We use a transformer's token-level prediction tasks (e.g., BERT \cite{devlin2018bert}) to predict words that may serve as a substitute for the given instance.
Let $S = w_1w_2\cdots w_m$ be a sentence of $n$ words with $w_i$ being the $i$-th word (a.k.a. token) and $S[i..j]$ the substring
$w_i \cdots w_j$. 
Let $A = S[i..j]$ be an answer key and 
$U = S[p..q]$ with $i\leq p \leq q \leq j$ 
an instance of $A$. 
We mask $U$ and denote the masked sentence by $$S_U = S[1..p-1][\text{MASK}]S[q+1..n],$$
and
predict words in the place of $U$ according to the surrounding context,
with conditional probabilities in descending order.

Predicted words and sibling entries of $U$ obtained from WordNet are initial instance-level distractor candidates (IDCs) for $U$.
Denote by ${\cal Y}_U$ the set of these IDCs (where there is no confusion, the subscript may be omitted). 
Let $Y_U \in {\cal Y}_U$.

\paragraph{Sense disambiguation} \label{sec:disambiguation.}

We use ESCHER \cite{barba2021esc} to determine the sense of $U$ with respect to $S$.
ESCHER is a transformer-based model for extractive sense comprehension.
It takes a sentence segmented into a sequence of instances as input and computes the sense of each instance.
Outperforming other sense disambiguation models, 
ESCHER achieves an F1 score of 83.9\% for noun instances and an F1 score of 69.3\% for verb instances with manually specified instances
over a combined dataset called SemEval, which is a union of SemEval-2007 (SE07), Senseval-2 (SE2), Senseval-3 (SE3), SemEval-2013 (SE13), and SemEval-2015 (SE15) \cite{SE07,SE15,SE13,Senseval3,Senseval2}. 

We perform sense disambiguation for $Y_U$ in the same way by replacing $U$ with $Y_U$ in $S$ as an input to ESCHER.

\paragraph{Instance and IDC information extraction} 

In addition to determining the synsets of $U$ and $Y_U$ with respect to $S$, we would also want to obtain, whenever possible, their POS tags, NER (named entity recognition) tags, lexical labels, and $U$'s inherited hypernym synsets. Let $\sigma$ be a synset. The inherited hypernym synsets of $\sigma$ comprise ancestry synsets of $\sigma$.
%
In particular, we obtain POS and NER tags using NLP tools such as spaCy taggers.
We use ESCHER to identify its synset $\sigma$, then search for the lexical label of $\sigma$.
We query WordNet for inherited hypernym synsets of $\sigma$. 
See Table \ref{table:type-1} for samples of such information.

Next, we replace [MASK] with $Y$ in 
the sentence $S$ to compute for $Y$ its POS tag and NER tag using spaCy, its synset using ESCHER and then
its lexical tag, and its computed synset's inherited hypernym synsets from WordNet. 

\section{Answer-key Distractor Generator}
\label{sec:ADG}

\begin{figure}[hbtp]
\centering
\includegraphics[width=\columnwidth]{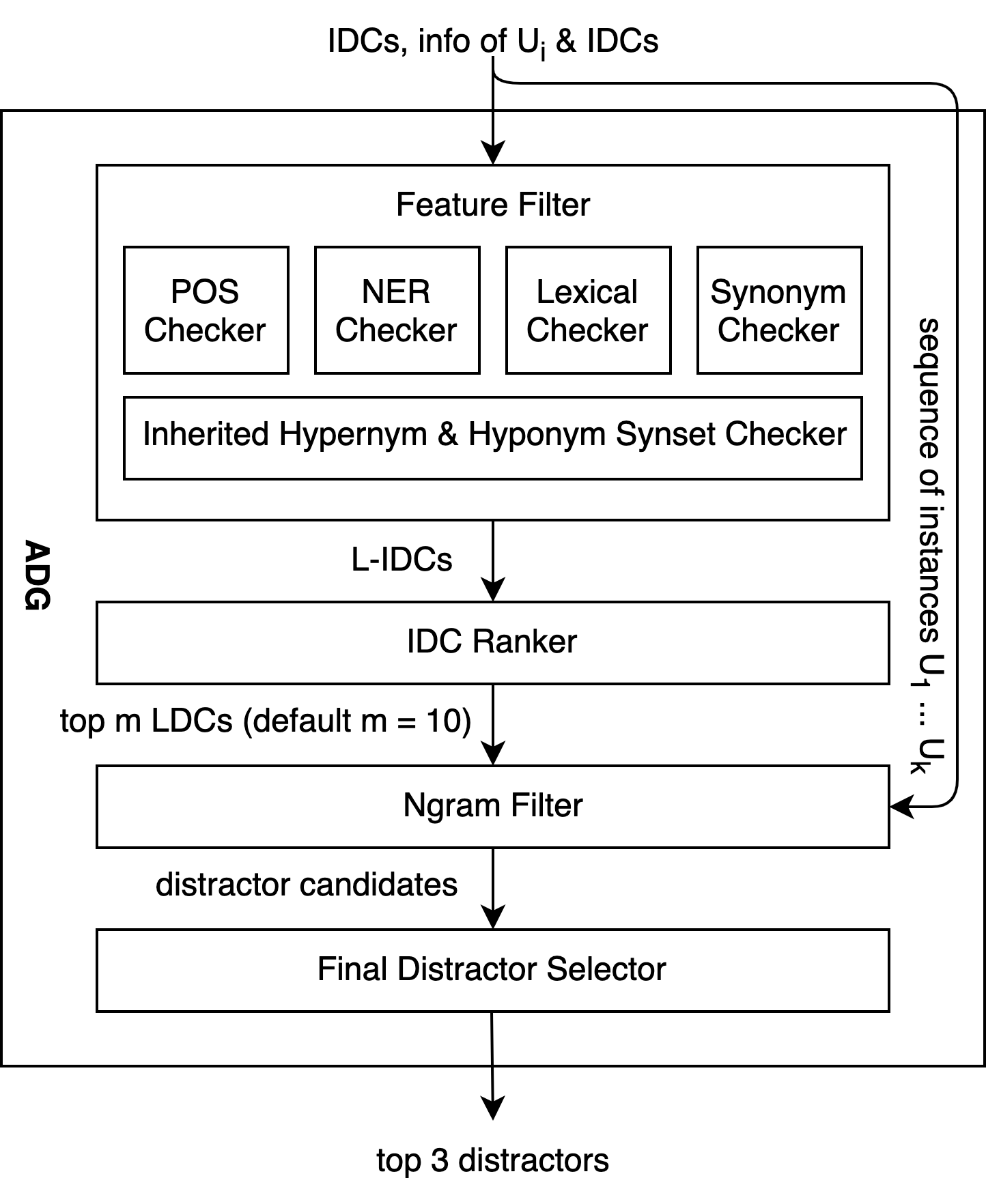}
\caption{ADG sub-components and data flow}
\label{fig:ADG}
\end{figure}

ADG consists of four sub-components (see Fig. \ref{fig:ADG}): (1) Feature Filter. (2) IDC Ranker. (3) Ngram Filter. (4) Final Distractor Selector.

\subsection{Feature filter} \label{sec:filter-1}

The feature filter consists of a sequence of five checkers: 
POS, NER, Lexical, Synonym, and Inherited Hypernym \& Hyponym Synset (IHHS).

\paragraph{POS and NER checkers}
Remove $Y$ if the POS tag or NER tag of $Y$ differs from
the corresponding tag of $U$.
The requirement that $Y$ and $U$ should have the same part of speech follows
the guidelines for writing up English vocabulary questions \cite{heaton1988writing}. 
We extend this requirement to named entities.

\paragraph{Lexical checker}
Analyzing unigram distractors in the SciQ database \cite{welbl2017crowdsourcing}, 
we find that in most cases, human-written distractors and the corresponding answer keys have the same lexical labels in WordNet, which happens about 94\% of the time. Thus, we use WordNet's lexical groupings of synsets to reduce the range for finding IDCs.
Synsets in WordNet 
are grouped into 45 lexicographer files, 
which are divided into 26 noun files with
lexical labels 
from \textsl{noun.Tops} (unique beginner for nouns) to \textsl{noun.time} (nouns denoting time and temporal relations), 15 verb files with lexical labels from \textsl{verb.body} (verbs of grooming, dressing and bodily care) to \textsl{verb.weather} (verbs of raining, snowing, thawing, thundering), three adjective files, and one adverb file. Under this lexical grouping, 
each synset belongs to exactly one lexical group.
For example, the eight synsets of \textsl{dog} are divided into
five lexicographer files with the following lexical labels: \textsl{noun.animal}, \textsl{noun.artifact}, \textsl{noun.food}, \textsl{noun.person}, and \textsl{verb.motion}.

Denote by $\ell_{U}$ the lexical label of $U$. Let
$L_Y$ be the set of lexical labels for the lexicographer files containing
the synsets of $Y$.
Furthermore, let $\ell_{Y}$ be the lexical label of $Y$ obtained by replacing $U$ with $Y$ in sentence $S$.
The lexical checker removes $Y$ if $\ell_U \not= \ell_Y$. 

We may also use a relaxed version of lexical checker, for human-written distractors 
may have different lexical labels from that of the instance. For example,
in the sentence ``\textsl{The average human body contains 5,830 g of blood}" with the answer key \textsl{blood}
whose lexical label is \textsl{noun.body}, the following distractors are provided by human writers: 
(1) \textsl{muscle} (\textsl{noun.body}), (2) \textsl{bacteria} (\textsl{noun.animal}), and (3)
\textsl{water} (\textsl{noun.substance}), where lexical labels with respect to the sentence are parenthesized.
The relaxed lexical checker checks if 
$\ell_{U} \in L_Y$. If no,  remove $Y$.
Nevertheless, in our experiments, we will use the stronger version.

\paragraph{Synonym checker}
Remove $Y$ if $Y$ is a synonym of $U$, i.e, if the synset of $Y$ is the same as that of $U$. 

\paragraph{IHHS checker}
Finally, let $H_U$ be the set of $U$'s inherited hypernym and hyponym synsets and 
$S_Y$ the set of $Y$'s synsets.
Remove $Y$
if $H_U \cap S_Y = \emptyset$.

The remaining IDCs that survive the feature filter are referred to as L-IDCs.
Table \ref{table:type-1} depicts examples of using the feature filter to remove unfit IDCs and 
samples of L-IDCs.

\begin{table}[tbp]
\caption{\textbf{Sentence}: \textsl{Virchow was the first scientist to discover that leukemia is caused by rapid production of abnormal white blood cells.} \textbf{Answer keys}: \textsl{(1) first scientist; (2) abnormal white blood cells}. \textbf{Instances}: \textsl{(1) first, scientist; (2) abnormal, white blood cells}. \textbf{Checking order}: POS, NER, Lexical, Synonym, IHHS, where
LL stands for lexical label, ORD for ordinal, and Synm for Synonym. The table shows all 5 checkers in action and two
L-IDCs (in bold), where \textsl{white blood cells} and \textsl{red blood cells} are phrases.}
\label{table:type-1}
\includegraphics[width=\columnwidth]{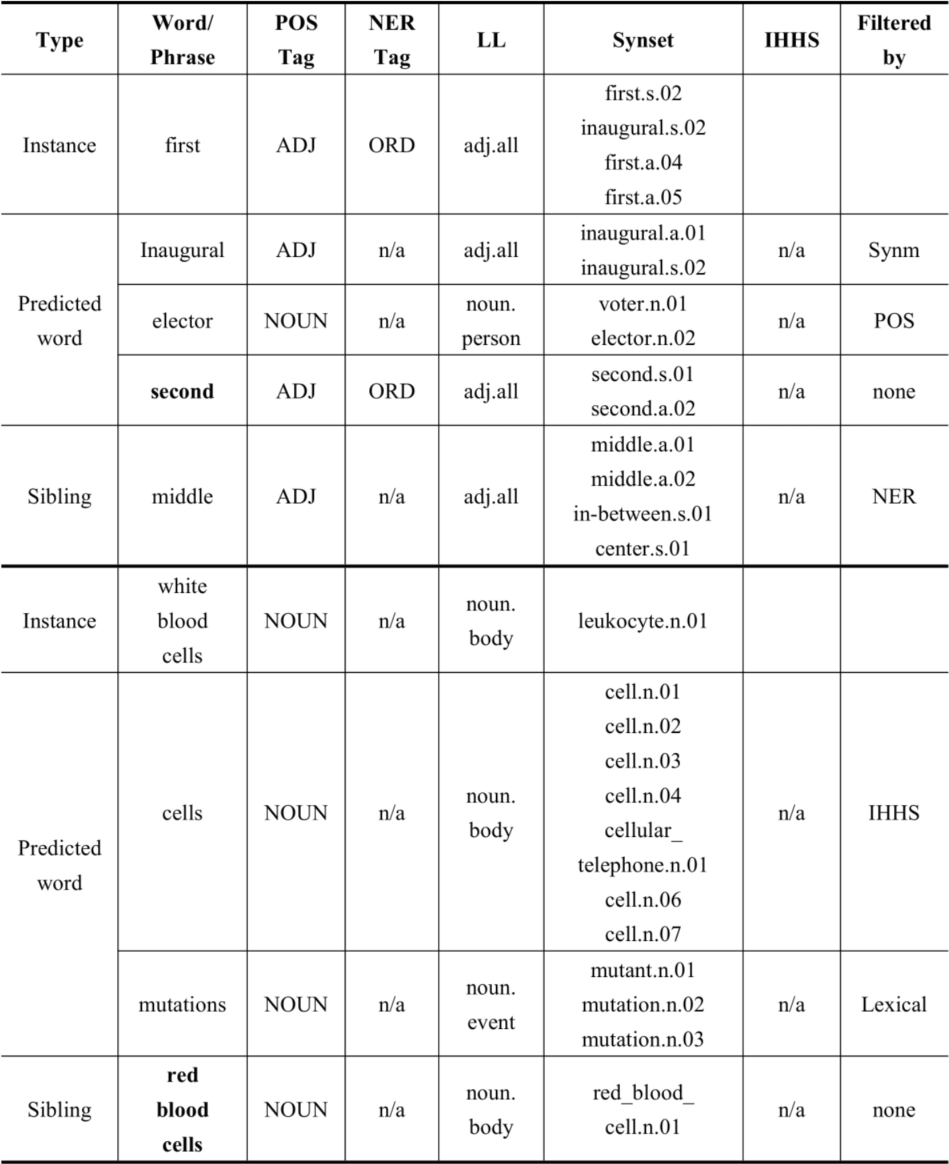}
\end{table}

\subsection{IDC Ranker} 
\label{sec:ranking}

Let $C$ be an L-IDC.
We use the contextual embedding similarity of  ${U}$ and $C$
as the baseline ranking of $C$, denoted by E$_S(U,C)$. We define 
WordNet synset similarity reward score W$_S(U,C)$ and prediction reward score P$_S(C)$ 
to adjust the baseline ranking score. 
The ranking of $C$, denoted by $\text{R}_S({U}, C)$, 
is defined to be the product of these metrics:
$$\text{R}_S(C) = \text{E}_S(U,C) \cdot \text{W}_S(U,C) \cdot \text{P}_S(C).$$

Let
$\bm{e}_S({U})$ denote the contextual embedding of ${U}$ with respect to $S$ obtained from a text-to-text transformer such as BERT, and define $\bm{e}_S(C)$ similarly. Denote by
$\text{E}_S({U},C)$
the cosine similarity of $\bm{e}_S({U})$ and $\bm{e}_S(C)$ as follows, with $\bm{\cdot}$ representing the dot product and $\Vert \bm{x}\Vert_2$ the Euclidean norm of $\bm{x}$:
\begin{equation}
\text{E}_S({U},C) = \frac{\bm{e}_S({U}) \bm{\cdot} \bm{e}_S(C))}{\Vert{\bm{e}_S({U})}\Vert_2 \cdot 
\Vert{\bm{e}_S(C)}\Vert_2}
\end{equation}

Denote by $\theta_{S,{U}}$ the synset of ${U}$ with respect to $S$, and $\ell_{U}$ the lexical label for
$\theta_{S,{U}}$.
Let $\Theta_{C,\ell_{U}}$ denote the set of synsets of $C$ with the same lexical label $\ell_{U}$.

Let $\sigma_1$ and $\sigma_2$ be two synsets. We define the semantic similarity of definitions of 
$\sigma_1$ and $\sigma_2$, denoted by SD$(\sigma_1, \sigma_2)$, to measure their relatedness 
by the cosine similarity of the BERT embeddings of the definitions of $\sigma_1$ and $\sigma_2$.
Note that $0 \leq \text{SD}(\sigma_1,\sigma_2) \leq 1$. We reward the baseline score $\text{E}_S({U},C)$ 
with
 as follows:
\begin{equation}
\text{W}_S({U},C) = 1 + [\max_{\theta \in \Theta_{C,\ell_{U}}} \text{SD}(\theta_{S,{U}}, \theta)]^\alpha,
\end{equation}
where $\alpha >0$ is a weight parameter whose value is to be trained via grid search. 

Predicted candidates come with conditional probabilities, 
which often descend too fast. It is common that the conditional probabilities of first two predicted words 
add up to over 98\% of the total weight, and the conditional probabilities for the rest of the words descend quickly  to less than \num{1e-5}. Yet it is possible that candidates good as distactors have very small conditional probabilities, and we would want to retain them.

\begin{figure}[tbp] 
	\centering 
	\includegraphics[width=\columnwidth]{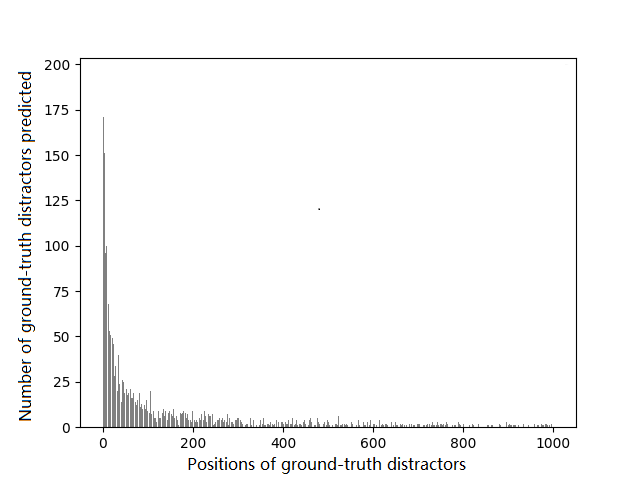} 
	\caption{Distribution of ground-truth distractors predicted} 
	\label{figure:position}
\end{figure}

For this purpose we use the ranking position of a candidate in the descending list of the predicted words according to conditional probability rather than the probability value itself. We show that, through numerical analysis, the distribution of 
BERT-predicted words that are also 
ground-truth distractors in the SciQ dataset 
follows the power law with the majority of weights on top-ranked words and a long tail (see Fig. \ref{figure:position}). 
Thus, looking at a few top predicted words would be sufficient for producing three distractors. We define the prediction reward score below: 
\begin{equation}
\text{P}_S(C) = \left\{
\begin{array}{ll}
1+{\beta}/{p_i}, &\mbox{if $C$ is predicted,} \\
1, &\mbox{otherwise,}
\end{array}
\right.
\end{equation}
where $p_i$ is the $i$-{th} position of the predicted candidate $C$, and $\beta \geq 0$ is a weight parameter 
whose value is to be trained via grid search.

\subsection{Ngram filter} \label{sec:filter-2}

For each instance ${U}$ we consider top $m$ $C$'s according to $\text{R}_S(C)$ (e.g., let $m=10$).
Let the answer key be a sequence of instances ${U}_1 \cdots {U}_k$. 
To form a distractor candidate, select an instance ${U}_i$ with $i \in [1,k]$. Let 
$P_i = \{C_{i_1}, \ldots, C_{i_m}\}$ be the set of L-IDCs for ${U}_i$. We 
replace ${U}_i$ with a $C_{i_j} \in P_i$ for one or more instances combinatorially to produce
a set of new phrases. This means that
we look at all possible ways of replacing instances with L-IDCs. For example, suppose that 
[A, B] is an answer key with two instances A and B. Suppose that C and D are IDCs for A and B, respectively, then [A, D], [C, B], and [C, D] are all combinations.
To determine if a new phrase conforms to human writing,
we use Google's Ngram Viewer \cite{google-ngram-viewer-2012} to check if the new phrase returns a result. If yes, then it confirms that the new phrase
has been written by a human writer and so we keep it as a distractor candidate. If no, we remove it. 

\subsection{Final distractor selector} 
\label{sec:distractors}

Let $D$ be a distractor candidate
that survives the Ngram filter. 
We define the ranking of $D$ with respect to
$A$ under $S$, denoted by $\text{R}_S(A, D)$, to be the 
cosine similarity of the
contextual embeddings $\bm{e}_S(A)$ and $\bm{e}_S(D)$.
FDS outputs the top $n$ distractor candidates as distractors for $A$ with $n = 3$ by default.

\subsection{Remark on distractor length}
Existing methods typically require that the length of a distractor be the same as that of the
answer key to ensure that both have the same sequence of POS tags of words.
We note that, however,
an appropriate distractor may or may not be of the same length of the answer key.
Our method can generate distractors of different lengths, 
as the length of a sibling entry of an instance may differ.
If the underlying text-to-text transformer can predict phrases -- this can be done, e.g., by retraining a BERT model on datasets segmented according to phrases,
we may also obtain distractor candidates of different lengths.
Moreover, we may remove some or all premodifying instances at random to reduce the length of the answer key. 

\section{Evaluations} \label{sec:evaluations}

We implement CQG, run numerical experiments, 
and carry out
ablation results on different ranking methods. We use BERT as text-to-text transformer and WordNet version 3.0 to train CQG.
%

\subsection{Datasets}

We use two datasets for evaluating CQG: U-SciQ and M-SciQ, where
U-SciQ consists of cloze questions with only unigram answer keys and
M-SciQ consists of cloze questions with only multigram answer keys.
Both datasets are separated from
the same dataset provided by Ren et al \cite{ren2021knowledge}, 
denoted by SciQ$^+$, 
which
was compiled from SciQ \cite{welbl2017crowdsourcing}, MCQL \cite{liang2018distractor}, 
AI2 Science Questions \cite{Clark2015ElementarySS}, and vocabulary MCQs crawled from the Internet,
with unigram and multigram answer keys.
In particular, U-SciQ is obtained from SciQ$^+$ by removing entries with multigram answer keys and converting interrogative sentences to cloze questions so that each entry
consists of a stem, an answer key, and three distractors.
Likewise, M-SciQ.
is obtained from SciQ$^+$ in the same way as U-SciQ except that this time,
entries with unigram answer keys are removed. 
U-SciQ consists of a total of 2,880 cloze questions, 
of which 1,176 are from SciQ, 300 from MCQL, 275 from AI2, and the rest from the Internet. 
M-SciQ consists of a total of 252 cloze questions, of which 210 have bigram answer keys, 39 have trigram answer keys, 
 and three have answer keys of length greater than 3.

\subsection{Model training}
We train four variants of CQG models: CQG-E is CQG using the baseline ranking method for L-IDCs, 
CQG-E/W, CQG-E/P, and 
CQG-E/W/P are CQG using E and W, E and P, and all ranking metrics, respectively. Namely, CQG is CQG-E/W/P.
In particular, we divide U-SciQ with a 3-1 split and apply grid search with cross validation  to find the best values of the parameters that achieve the highest F1 scores, 
with an increment of 0.1. The trained values are as follows:
$\alpha = 5.2$ for CQG-E/W, 
$\beta = 1.1$ for CQG-E/P, and $\alpha = 20.5$, $\beta = 1.1$ for CQG-E/W/P.

\textbf{Remark}.
Our evaluations use the stronger version of lexical checker, which removes candidates with lexical labels different from that of the answer key. However, good distractors may have different lexical labels. For example, in stem “A bee will sometimes do a dance to tell other bees in the hive where to find **blank**.” The answer key is \textsl{food} and the ground-truth distractors are \textsl{honey}, \textsl{enemies}, and \textsl{water}. The distractor \textsl{enemies} has a different lexical label from \textsl{food}. The relaxed version allows us to keep such distractors.
If we use the relaxed lexical checker, we would need to add a penalty score 
as follows by multiplying a lexical penalty score L$_S(U,C)$ to the rank R$_S(U,C)$, with L$_S(U,C)$:
\begin{equation}
\text{L}_S({U},C)=
\begin{cases}
\gamma, &\text{if $\ell_S(C) \not= \ell_S({U})$}, \\
1,  &\text{otherwise},
\end{cases}
\end{equation}
where $\gamma \in [0,1]$ is to be determined via grid search.

\subsection{Intrinsic evaluation}

\paragraph{Over U-SciQ}
We apply the same evaluation metrics used by Ren et al \cite{ren2021knowledge} 
to carry out intrinsic evaluations on U-SciQ, 
averaging over top 3 generated distractors with the 3 ground-truth distractors pairwise unless otherwise stated. These metrics include F1 score (F1), precision (P@1, P), recall (R),
mean reciprocal rank (MRR), normalized discounted cumulative gain (NDCG@10), and Word2Vec Semantic Similarity (WSS), where @$k$ means averaging over top $k$ distractors against the three ground truth distractors pairwise, and
WSS is the average cosine similarity of the top three generated distractors and ground truth distractors with 
Word2Vec embeddings trained on a Wikipedia dump. However, WSS does not represent words of multiple senses, and so we also use Contextual Semantic Similarity (CSS), which is the average cosine similarity of the top three generated distractors and ground truth distractors with embeddings generated by BERT for the underlying sentences.

Table \ref{table:U-SciQ} provides comparisons with SOTA results \cite{ren2021knowledge}, where
WN-CSG+DS and PB-CSG+DS denote, respectively, WordNet CSG+DS and Probase CSG+DS mentioned in 
Section \ref{sec:related-work}. In particular, PB-CSG+DS provides the SOTA results and WN-CSG+DS uses WordNet hypernyms for comparisons, and their results, except those for CSS/B and CSS/L,
are extracted from the reference \cite{ren2021knowledge}. Note that the SOTA results for WSS are fractional numbers and
we use the percentage for consistence of the rest of the values.
Finally, CSS/B and CSS/L denote the cosine similarity of contextual embeddings generated by, respectively, BERT-Base and BERT-Large.

It is evident that each CQG variant 
under F1, MRR, NDCG@10, and SS measures, and 
outperforms the SOTA results under all metrics except the recall R.
Moreover, each CQG variant is significantly better than the SOTA results under 
MRR and NDCG@10.
This indicates that CQG generates more
ground truth distractors that are also ranked high. We further note that
SOTA results are skewed to recall while CQGs are much better balanced between precision and recall, with higher F1 scores than those of SOTA results. Lower recall means that the generated distractors (even though they are good) contain less ground-truth distractors given in the dataset. This intrinsic evaluation only computes how much generated distractors overlap with the ground-truth distractors. Finally, we can see that CQG-E/W/P outperforms all other variants of CQG.

\begin{table}[tbp]
		\centering
		\caption{End-to-end machine evaluations on U-SciQ}
		\label{table:U-SciQ}
		\includegraphics[width=\columnwidth]{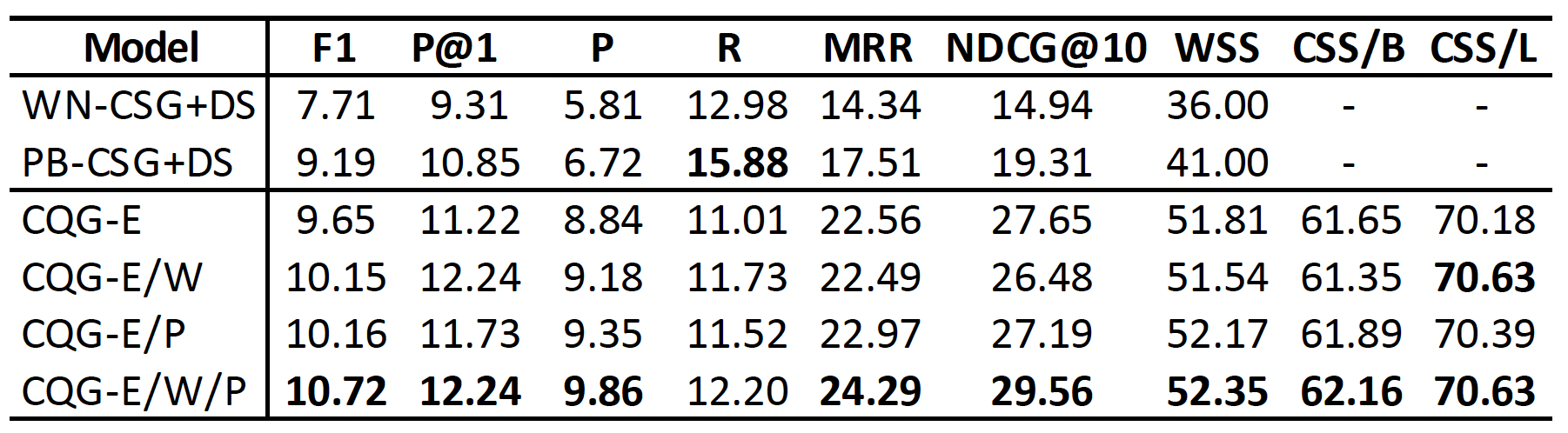}
\end{table}

\paragraph{Over M-SciQ}
For multigram answer keys and distractors, the performance measures used for unigrams no longer
work well. Instead, we use the
BLEU (B-1, B-2, B-3, B-4) and ROUGE (R-1, R-2, R-L) 
metrics to measure the average quality of the top 3 generated distractors against the ground truth.
B-$n$ evaluates average n-gram precision on ground truth distractors, and
R-1 and R-2 is the recall of unigrams and bigrams while 
R-L is the recall of longest common subsequences. 
Table \ref{table:M-SciQ} 
depicts the evaluation results.

We see that, while the BLEU-1 and ROUGE-L scores are relatively satisfactory, 
the BLEU-2 and BLEU-3 
scores are not.

\begin{table}[tbp]
		\centering
		\caption{End-to-end machine evaluations on M-SciQ}
		\label{table:M-SciQ}
		\includegraphics[width=\columnwidth]{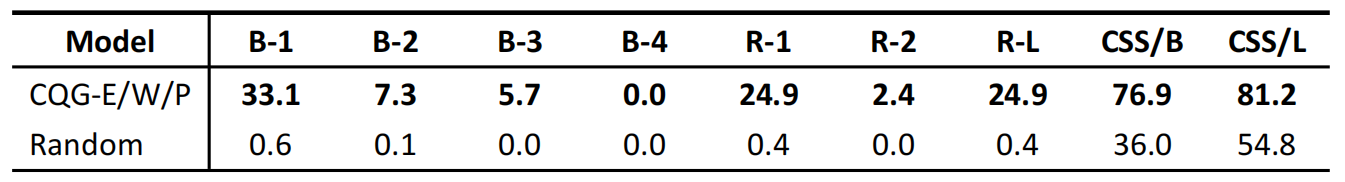}
\end{table}

It turns out in this case, R-L is the same as R-1.
This happens because there are often multiple choices of appropriate distractors and BLEU and ROUGE calculate the frequency of the same words appearing in the ground truth and the candidates. Thus, semantic similarity would be a better measure for multigrams. We see that CQG produces distractors that match semantically the ground truth
distractors with pairwise comparisons over 81\% of the time. Here by ``Random" it means random selection of L-IDCs without rankings. It makes sense that CSS/L score of random selection is larger than 50\% because L-IDCs are already contextually and semantically related to the corresponding answer keys.

Table \ref{table:concrete} depicts an example for sentence \textsl{mushrooms gain their energy from decomposing dead organisms}
with the answer key being \textsl{decomposing dead organisms}. We can see that even though
there is only one word common in both ground truth distractors and generated distractors,
where the union of both sets of distractors has 13 different words with one word \textsl{dead} in common, the CSS/L score is as high as 84.7\%.

\begin{table}[btp]
\caption{Sentence: Mushrooms gain their energy from decomposing dead organisms. Answer key: \textsl{decomposing dead organisms}}
\label{table:concrete}
\includegraphics[width=\columnwidth]{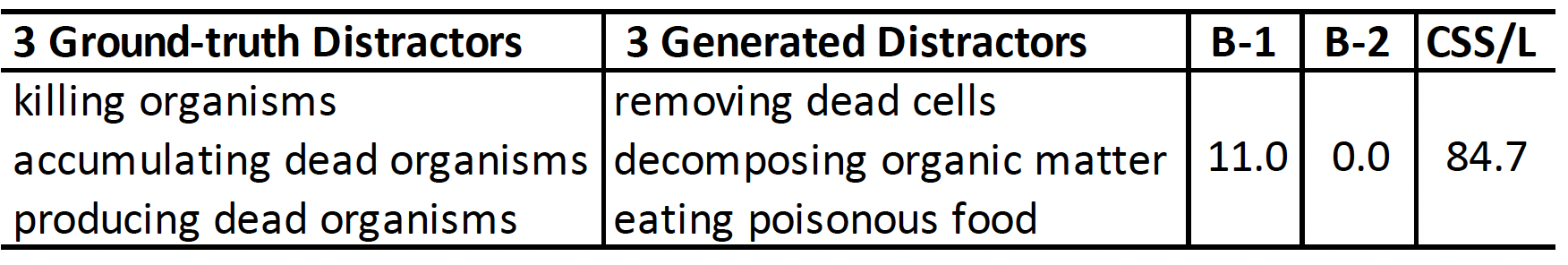}
\end{table}

In the previous example on sentence ``Leukemia is caused by the rapid production of abnormal white blood cells" with answer key = \textsl{abnormal white blood cells}, CQG outputs \textsl{abnormal red blood cells, defective genes, normal serum} as distractors. 

\subsection{Extrinsic evaluation} 

Three human judges evaluate distractors' reliability and plausibility for cloze questions generated over U-SciQ and M-SciQ with the following guidelines: A distractor receives a reliability score of 1 if 
placing the distractor in the blank space of the stem
produces a contextually fit and grammatically correct yet logically wrong sentence, and 0 otherwise.
For example, if the stem is about biology but a distractor is in the domain of physics, then the 
the reliability score of this distractor should be 0 even if the new sentence is grammatically correct, because it is contextually unfit.
The judges assess the plausibility of a distractor on a 3-point scale: a distractor receives 0 points if it is obviously wrong, 2 points if it is sufficiently confusing as to which is the absolute correct answer,
and 1 point if it is somewhat confusing. 

The judges are presented with the evaluation
dataset of U-SciQ and M-SciQ, where each data point consists of a stem, the corresponding answer keys and
3 ground-truth distractors, mixing with 3 CQG-generated distractors, but
the judges do not know which distractors are ground truth and which are CQG generated.
In so doing we hope to achieve unbiased judging. All judges have written exam questions with at least three years of teaching experience.
We compare judges' scores on CQG-generated distractors against ground-truth distractors.
The results are given in Table \ref{table:human}.

\begin{table}[tbp]
\centering
\caption{Blind evaluations by three judges who do not know which distractors are CQG generated and which are ground truth,
where CQG-R, GT-R, CQG-P, and GT-P stand for, respectively, CQG Reliability, Ground-truth Reliability,
CQG Plausibility, and Ground-truth Plausibility. Moreover, Judge-Avg and Stdev stand for, respectively, average and standard deviation of all judges.} 
\label{table:human}
\begin{tabular}{l|l|c|c|c|c}
\hline
\textbf{Dataset}&\textbf{Judge} &\textbf{CQG-R}	& \textbf{GT-R} &	\textbf{CQG-P}  & \textbf{GT-P} \\\hline
\multirow{ 5}{*}{U-SciQ} 
&Judge 1 & 	0.9639	&0.9744	&1.8041	&1.8460 \\
&Judge 2 &	0.9676	&0.8693	&1.6325	&1.1987 \\
&Judge 3	&   0.9761	&0.9829	&1.8176	&1.7431 \\\cline{2-6}
&Avg	&   \textbf{0.9692}	&0.9422	&\textbf{1.7514}	&1.5960 \\
&Stdev	&   0.0063	&0.0632	&0.1032	&0.3478 \\\hline
\multirow{ 5}{*}{M-SciQ}
&Judge 1	&   0.9877	&0.9902	&1.8151	&1.8469 \\
&Judge 2	&   0.9830	&0.9496	&1.6834	&1.5035 \\
&Judge 3	&   0.9878	&0.9709	&1.8589	&1.8062 \\\cline{2-6}
&Avg	&   \textbf{0.9862}	&0.9702	&\textbf{1.7858}	&1.7189 \\
&Stdev	&   0.0027	&0.0203	&0.0913	&0.1876 \\\hline
\end{tabular}
\end{table}

It can be seen that judges view the qualities of CQG generated distractors higher than
those of ground-truth distractors on both reliability and plausibility. Moreover, the evaluation results of CQG-generated distractors are consistent
between judges, while the evaluation results of ground-truth distractors are fluctuating. 
To understand why this
would be the case, we investigate each entry and its evaluation results. For example, in the following stem
\textsl{the name for a biologist who studies fungi is **blank**} with the answer key \textsl{mycologists},
the CQG-generated distractors are \textsl{zoologist}, 
\textsl{chemist}, 
and \textsl{anthropologist}, 
while
the ground-truth disasters are	
\textsl{egyptologists}, 
\textsl{musicologists}, 
and \textsl{oncologists}. 
It is evident that all three CQG-generated distractors are contextually fit, while 
the ground-truth distractors \textsl{egyptologists}	and \textsl{musicologists} do not match the context of biologists. 
In the following stem	
\textsl{in mammals, four specialized types of **blank** serve to cut, tear, and grind food} with
the answer key \textsl{teeth}, 
the CQG-generated distractors are \textsl{fingernails}, 
\textsl{nails}, 
and \textsl{claws}, 
while the ground-truth distractors
are \textsl{scales}, 
\textsl{plates}, 
and \textsl{spines}. 
Clearly, plates and spines cannot cut, tear, or grind food.	
Such contextually unfit distractors do occur more often in ground-truth distractors, which contribute to the lower
average reliability score of ground-truth distractors.

For another example, in the stem \textsl{energy that is stored in the connections ***blank*** in a chemical substance} with the answer key \textsl{between atoms}, the CQG-generated distractors are 
\textsl{between groups}, 
\textsl{between gases}, 
and \textsl{within hydrogen}	
while the ground-truth distractors are
\textsl{on the surface}, 
\textsl{in molecules},	
and \textsl{inside atoms}. 
It is evident that all ground-truth distractors are problematic to succeed the phrase \textsl{in the connections}, while only one CQG-generated distractor \textsl{within hydrogen} has this problem.
Thus, these four distractors would all receive 0 for reliability and 0 for plausibility.Even when there is no grammatical issue, ground-truth distractors seem to be more often to fail the plausibility test.

\section{Conclusions and Final Remarks} \label{sec:conclusions}
CQG generates cloze questions of high quality and surpasses the state-of-the-art results. Techniques used in CQG can also be used to generate distractors for multiple-choice questions by forming interrogative sentences using transformers based on an answer key and surrounding declarative statements \cite{zhang-zhang-sun-wang2022}. To facilitate further research, we have published the CQG API, datasets, and assessment results by judges on \url{https://github.com/clozeQ/A-Generative-Method-for-Producing-Distractors-for-Cloze-Questions}.


The qualities of distractors generated by CQG depend on a number of factors, including sense disambiguation and WordNet taxonomies. As better sense disambiguation algorithms are developed and more words and phrases are added to WordNet, CQG is expected to generate distractors with higher qualities. 


The dataset SciQ$^+$ consists of stems converted from interrogative sentences, where a number of conversions are erroneous. For example, the stem \textsl{**blank** of devices scientists use to determine wind speed} is likely converted from \textsl{what kind of device do scientists use to determine wind speed}, with \textsl{anemometer} being the answer key. For this interrogative-sentence-like stem, our algorithm generates the following distractors: \textsl{type, tool, mechanism}, which are clearly not reliable. Using the corrected stem \textsl{scientists use **blank** to determine wind speed}, our algorithm generates the following appropriate distractors: \textsl{GPS, vacuum gage, binoculars}. We have posted on the aforementioned github link a cleaned version of the dataset to ensure that each stem is in the appropriate declarative form for future studies. 

\bibliographystyle{IEEEtran}
\bibliography{reference}

\begin{thebibliography}{10}
\providecommand{\url}[1]{#1}
\csname url@samestyle\endcsname
\providecommand{\newblock}{\relax}
\providecommand{\bibinfo}[2]{#2}
\providecommand{\BIBentrySTDinterwordspacing}{\spaceskip=0pt\relax}
\providecommand{\BIBentryALTinterwordstretchfactor}{4}
\providecommand{\BIBentryALTinterwordspacing}{\spaceskip=\fontdimen2\font plus
\BIBentryALTinterwordstretchfactor\fontdimen3\font minus \fontdimen4\font\relax}
\providecommand{\BIBforeignlanguage}[2]{{%
\expandafter\ifx\csname l@#1\endcsname\relax
\typeout{** WARNING: IEEEtran.bst: No hyphenation pattern has been}%
\typeout{** loaded for the language `#1'. Using the pattern for}%
\typeout{** the default language instead.}%
\else
\language=\csname l@#1\endcsname
\fi
#2}}
\providecommand{\BIBdecl}{\relax}
\BIBdecl

\bibitem{miller1998wordnet}
G.~A. Miller, \emph{WordNet: An electronic lexical database}.\hskip 1em plus 0.5em minus 0.4em\relax MIT press, 1998.

\bibitem{ren2021knowledge}
S.~Ren and K.~Q. Zhu, ``Knowledge-driven distractor generation for cloze-style multiple choice questions,'' in \emph{Proc. of the AAAI Conference on Artificial Intelligence}, vol.~35, no.~5, 2021, pp. 4339--4347.

\bibitem{sumita2005measuring}
E.~Sumita, F.~Sugaya, and S.~Yamamoto, ``Measuring non-native speakers’ proficiency of english by using a test with automatically-generated fill-in-the-blank questions,'' in \emph{Proc. of the second workshop on Building Educational Applications Using NLP}, 2005, pp. 61--68.

\bibitem{mitkov2009semantic}
R.~Mitkov, A.~Varga, L.~Rello \emph{et~al.}, ``Semantic similarity of distractors in multiple-choice tests: extrinsic evaluation,'' in \emph{Proc. of the workshop on geometrical models of natural language semantics}, 2009, pp. 49--56.

\bibitem{smith2010gap}
S.~Smith, P.~Avinesh, and A.~Kilgarriff, ``Gap-fill tests for language learners: Corpus-driven item generation,'' in \emph{Proc. of ICON: 8th International Conference on Natural Language Processing}, 2010, pp. 1--6.

\bibitem{jiang2017distractor}
S.~Jiang and J.~S. Lee, ``Distractor generation for chinese fill-in-the-blank items,'' in \emph{Proc. of the 12th Workshop on Innovative Use of NLP for Building Educational Applications}, 2017, pp. 143--148.

\bibitem{wu2012probase}
W.~Wu, H.~Li, H.~Wang, and K.~Q. Zhu, ``Probase: A probabilistic taxonomy for text understanding,'' in \emph{Proc. of the 2012 ACM SIGMOD international conference on management of data}, 2012, pp. 481--492.

\bibitem{blei2003latent}
D.~M. Blei, A.~Y. Ng, and M.~I. Jordan, ``Latent dirichlet allocation,'' \emph{Journal of machine Learning research}, vol.~3, no. Jan, pp. 993--1022, 2003.

\bibitem{susanti2018automatic}
Y.~Susanti, T.~Tokunaga, H.~Nishikawa, and H.~Obari, ``Automatic distractor generation for multiple-choice english vocabulary questions,'' \emph{Research and practice in technology enhanced learning}, vol.~13, no.~1, pp. 1--16, 2018.

\bibitem{knoop2013wordgap}
S.~Knoop and S.~Wilske, ``Wordgap-automatic generation of gap-filling vocabulary exercises for mobile learning,'' in \emph{Proc. of the Second Workshop on NLP for Computer-assisted Language Learning at NODALIDA 2013}, no. 086, 2013, pp. 39--47.

\bibitem{guo2016questimator}
Q.~Guo, C.~Kulkarni, A.~Kittur, J.~P. Bigham, and E.~Brunskill, ``Questimator: Generating knowledge assessments for arbitrary topics,'' in \emph{Proc. of the AAAI Twenty-Fifth International Joint Conference on Artificial Intelligence}, 2016.

\bibitem{brown2005automatic}
J.~Brown, G.~Frishkoff, and M.~Eskenazi, ``Automatic question generation for vocabulary assessment,'' in \emph{Proc. of Human Language Technology Conference and Conference on Empirical Methods in Natural Language Processing}, 2005, pp. 819--826.

\bibitem{mitkov2003computer}
R.~Mitkov \emph{et~al.}, ``Computer-aided generation of multiple-choice tests,'' in \emph{Proc. of the HLT-NAACL 03 workshop on Building educational applications using natural language processing}, 2003, pp. 17--22.

\bibitem{agarwal2011automatic}
M.~Agarwal and P.~Mannem, ``Automatic gap-fill question generation from text books,'' in \emph{Proc. of the sixth workshop on innovative use of NLP for building educational applications}, 2011, pp. 56--64.

\bibitem{pino2008selection}
J.~Pino, M.~Heilman, and M.~Eskenazi, ``A selection strategy to improve cloze question quality,'' in \emph{Proc. of the Workshop on Intelligent Tutoring Systems for Ill-Defined Domains. 9th International Conference on Intelligent Tutoring Systems, Montreal, Canada}, 2008, pp. 22--32.

\bibitem{mostow2012generating}
J.~Mostow and H.~Jang, ``Generating diagnostic multiple choice comprehension cloze questions,'' in \emph{Proc. of the Seventh Workshop on Building Educational Applications Using NLP}, 2012, pp. 136--146.

\bibitem{liang2017distractor}
C.~Liang, X.~Yang, D.~Wham, B.~Pursel, R.~Passonneaur, and C.~L. Giles, ``Distractor generation with generative adversarial nets for automatically creating fill-in-the-blank questions,'' in \emph{Proc. of the Knowledge Capture Conference}, 2017, pp. 1--4.

\bibitem{liang2018distractor}
C.~Liang, X.~Yang, N.~Dave, D.~Wham, B.~Pursel, and C.~L. Giles, ``Distractor generation for multiple choice questions using learning to rank,'' in \emph{Proc. of the thirteenth workshop on innovative use of NLP for building educational applications}, 2018, pp. 284--290.

\bibitem{spacy}
M.~Honnibal and I.~Montani, ``{spaCy 2}: Natural language understanding with {B}loom embeddings, convolutional neural networks and incremental parsing,'' 2017, to appear.

\bibitem{Zhang-Zhou-Wang2021}
H.~Zhang, Y.~Zhou, and J.~Wang, ``Contextual networks and unsupervised ranking of sentences,'' in \emph{2021 IEEE 33rd International Conference on Tools with Artificial Intelligence (ICTAI)}.\hskip 1em plus 0.5em minus 0.4em\relax IEEE, 2021, pp. 1126--1131.

\bibitem{devlin2018bert}
J.~Devlin, M.-W. Chang, K.~Lee, and K.~Toutanova, ``Bert: Pre-training of deep bidirectional transformers for language understanding,'' \emph{arXiv preprint arXiv:1810.04805}, 2018.

\bibitem{barba2021esc}
E.~Barba, T.~Pasini, and R.~Navigli, ``Esc: Redesigning wsd with extractive sense comprehension,'' in \emph{Proc. of the 2021 Conference of the North American Chapter of the Association for Computational Linguistics: Human Language Technologies}, 2021, pp. 4661--4672.

\bibitem{SE07}
S.~Pradhan, E.~Loper, D.~Dligach, and M.~Palmer, ``Semeval-2007 task-17: English lexical sample, srl and all words,'' in \emph{Proc. of the fourth international workshop on semantic evaluations (SemEval-2007)}, 2007, pp. 87--92.

\bibitem{SE15}
A.~Moro and R.~Navigli, ``Semeval-2015 task 13: Multilingual all-words sense disambiguation and entity linking,'' in \emph{Proc. of the 9th international workshop on semantic evaluation (SemEval 2015)}, 2015, pp. 288--297.

\bibitem{SE13}
R.~Navigli, D.~Jurgens, and D.~Vannella, ``Semeval-2013 task 12: Multilingual word sense disambiguation,'' in \emph{Proc. of the Seventh International Workshop on Semantic Evaluation (SemEval 2013)}, 2013, pp. 222--231.

\bibitem{Senseval3}
\BIBentryALTinterwordspacing
B.~Snyder and M.~Palmer, ``The {E}nglish all-words task,'' in \emph{Proc. of {SENSEVAL}-3, the Third International Workshop on the Evaluation of Systems for the Semantic Analysis of Text}.\hskip 1em plus 0.5em minus 0.4em\relax Barcelona, Spain: Association for Computational Linguistics, Jul. 2004, pp. 41--43. [Online]. Available: \url{https://aclanthology.org/W04-0811}
\BIBentrySTDinterwordspacing

\bibitem{Senseval2}
\BIBentryALTinterwordspacing
P.~Edmonds and S.~Cotton, ``{SENSEVAL}-2: Overview,'' in \emph{Proc. of {SENSEVAL}-2 Second International Workshop on Evaluating Word Sense Disambiguation Systems}.\hskip 1em plus 0.5em minus 0.4em\relax Toulouse, France: Association for Computational Linguistics, Jul. 2001, pp. 1--5. [Online]. Available: \url{https://aclanthology.org/S01-1001}
\BIBentrySTDinterwordspacing

\bibitem{heaton1988writing}
J.~B. Heaton, ``Writing {E}nglish language tests,'' 1988.

\bibitem{welbl2017crowdsourcing}
J.~Welbl, N.~F. Liu, and M.~Gardner, ``Crowdsourcing multiple choice science questions,'' \emph{arXiv preprint arXiv:1707.06209}, 2017.

\bibitem{google-ngram-viewer-2012}
\BIBentryALTinterwordspacing
Google, ``Google ngram viewer,'' 2012. [Online]. Available: \url{http://books.google.com/ngrams/datasets}
\BIBentrySTDinterwordspacing

\bibitem{Clark2015ElementarySS}
P.~Clark, ``Elementary school science and math tests as a driver for ai: Take the aristo challenge!'' in \emph{AAAI}, 2015.

\bibitem{zhang-zhang-sun-wang2022}
C.~Zhang, H.~Zhang, Y.~Sun, and J.~Wang, ``Downstream transformer generation of question-answer pairs with preprocessing and postprocessing pipelines,'' in \emph{Proc. of the 22nd ACM Symposium on Document Engineering}, 2022.

\end{thebibliography}

\end{document}